%% file: iclr2020_conference.tex
\documentclass{article} 
\usepackage{iclr2020_conference,times}

\input{math_commands.tex}

\usepackage{hyperref}
\usepackage{url}

\usepackage[utf8]{inputenc} 
\usepackage[T1]{fontenc}    
\usepackage{hyperref}       
\usepackage{url}            
\usepackage{booktabs}       
\usepackage{amsfonts}       
\usepackage{nicefrac}       
\usepackage{microtype}      

\usepackage{kotex}

\usepackage{graphicx}

\usepackage[english]{babel}

\usepackage{color}
\usepackage{xcolor}

\usepackage{algorithm}
\usepackage{algorithmicx}
\usepackage{algpseudocode}
\usepackage{hyperref}

\usepackage{amsmath,amssymb,amsfonts,amsthm,mathtools,bm}
\usepackage{enumitem}
\usepackage{multirow}
\usepackage{bm}
\usepackage{array}
\usepackage{color}
\usepackage{xcolor}
\usepackage{wrapfig}
\usepackage{multicol}
\usepackage{caption}
\usepackage{subcaption}
\usepackage{diagbox}
\usepackage{enumitem}

\usepackage{mathtools}
\usepackage{threeparttable}
\usepackage{standalone}
\usepackage{booktabs, dcolumn}

\title{
Efficient Click-Through Rate Prediction for Developing Countries via Tabular Learning
}

\author{Joonyoung Yi and Buru Chang \\
Hyperconnect, South Korea \\
\texttt{\{joonyoung.yi,buru.chang\}hpcnt.com} \\
}

\newcommand{\STAB}[1]{\begin{tabular}{@{}c@{}}#1\end{tabular}}

\begin{document}

\maketitle

\input{Sections/0_abstract}
\input{Sections/1_introduction}
\input{Sections/2_related_works}
\input{Sections/3_solution}
\input{Sections/4_experiments}
\input{Sections/5_conclusion}

\newpage
\bibliography{iclr2020_conference}
\bibliographystyle{iclr2020_conference}


\end{document}

%% file: math_commands.tex
\usepackage{amsmath,amsfonts,bm}

\def\eqref#1{equation~\ref{#1}}

\def\1{\bm{1}}

\DeclareMathAlphabet{\mathsfit}{\encodingdefault}{\sfdefault}{m}{sl}
\SetMathAlphabet{\mathsfit}{bold}{\encodingdefault}{\sfdefault}{bx}{n}



%% file: Sections/0_abstract.tex
\begin{abstract}\label{sec:0_abstract}

Despite the rapid growth of online advertisement in developing countries, existing highly over-parameterized Click-Through Rate (CTR) prediction models are difficult to be deployed due to the limited computing resources.
In this paper, by bridging the relationship between CTR prediction task and tabular learning, we present that tabular learning models are more efficient and effective in CTR prediction than over-parameterized CTR prediction models.
Extensive experiments on eight public CTR prediction datasets show that tabular learning models outperform twelve state-of-the-art CTR prediction models.
Furthermore, compared to over-parameterized CTR prediction models, tabular learning models can be fast trained without expensive computing resources including high-performance GPUs.
Finally, through an A/B test on an \textit{actual} online application, we show that tabular learning models improve not only offline performance but also the CTR of real users.
\end{abstract}

%% file: Sections/1_introduction.tex
\section{Introduction}\label{sec:1_introduction}

With the spread of mobile devices, the e-commerce market in developing countries is rapidly growing.
For example, the Indian e-commerce market is expected to grow to US\$ 200 billion by 2026 from US\$ 38.5 billion as of 2017\footnote{Reported by India Brand Equity Foundation (Nov 2020)}.
Accordingly, Click-Through Rate (CTR) prediction has become more important in online advertisements in developing countries as well as developed countries.
Regardless of industry and academia, highly over-parameterized CTR prediction models~\citep{lian2018xdeepfm, song2019autoint, cheng2020adaptive} have recently been proposed to improve performance using deep neural networks.
These CTR prediction models should train a considerable amount of parameters to deal with millions of input features.
Therefore, massive computing resources, including high-performance GPUs, are required to train the over-parameterized CTR prediction models.

However, these CTR prediction models are hard to be applied to real-world applications under limited computing resource scenarios of developing countries.
First, it is well known that real-world applications have a \textit{stale problem} that the performance of their CTR prediction models is severely degraded over time by many reasons including change of user/item pools~\citep{nasraoui2007performance, radinsky2012modeling, trevisiol2014cold} (See Figure~\ref{fig:online_ctr}).
Second, hyper-parameter tuning is essential for over-parameterized CTR prediction models because they tend to be sensitive to hyper-parameters~\citep{cheng2020adaptive}.
To alleviate these issues, solutions using daily training~\citep{koren2009collaborative, vartak2017meta, wang2019predicting}, meta-learning~\citep{lee2019melu,chen2019lambdaopt}, and reinforcement learning~\citep{jagerman2019people,shi2019pyrecgym} have been suggested.
However, since these solutions require sufficient computing resources, they are hard to be applied to real-world applications in developing countries.

In this paper, we bridge the relationship between CTR prediction and tabular learning~\citep{chen2015xgboost,ke2017lightgbm,dorogush2018catboost} to find a practical CTR prediction model that can be easily deployed in developing countries. 
In CTR prediction, conventional tabular learning models~\citep{ke2017lightgbm,dorogush2018catboost} have been neglected as baselines because input features in CTR prediction are mainly composed of highly sparse categorical features and it is difficult to be trained by the tabular learning models~\citep{ke2019deepgbm}.
Meanwhile, the recently proposed methods~\citep{ke2017lightgbm, dorogush2018catboost, ayria2020kfoldte} handling categorical features in tabular learning have dramatically improved the performance by slightly modifying primitive categorical feature encoders~\citep{fisher1958grouping, micci2001preprocessing}.
Borrowing tabular learning models with these methods to CTR prediction, we demonstrate that tabular learning models outperform existing CTR prediction models accompanying their cost-efficiency.

\paragraph{Our contributions.}
(1) We explicitly investigate the relationship between CTR prediction task and tabular learning, and suggest that tabular learning models could act as efficient baselines in CTR prediction.
(2) Extensive experiments show that tabular learning models outperform current CTR prediction models, accompanying their cost-efficiency.
We believe that our extensive experiment results of tabular learning models in CTR prediction can contribute to research communities related to CTR prediction.
(3) Finally, our online experiments including the A/B test on an active online application validate the effectiveness of tabular learning methods under limited computing resources.

%% file: Sections/2_related_works.tex
\section{Related Works}\label{sec:2_related_works}

\subsection{Click-Through Rate Prediction}
CTR prediction is to predict the probability of the user $u$ clicking on the item $v$.
The major difference with collaborative filtering~\citep{rendle2012bpr} is that CTR prediction utilizes additional side information about users and items as input features containing highly sparse categorical features.
Factorization Machines (FM)~\citep{rendle2010factorization} is the most representative CTR prediction model which considers the first and second-order feature interactions from input features, simultaneously. 
Recently, various CTR prediction models have been proposed to capture the high-order feature interactions via deep neural networks~\citet{cheng2016wide, he2017neural, guo2017deepfm, lian2018xdeepfm, qu2018product, song2019autoint, cheng2020adaptive}.
However, these CTR prediction models are too over-parameterized to be deployed under limited computing resource scenarios. 

\subsection{Tabular Learning}
Tabular learning refers to a learning methodology handling \textit{tabular heterogeneous data} as input. 
It is known that gradient boosting models based on decision trees~\citep{chen2015xgboost,ke2017lightgbm,dorogush2018catboost} show superior performance in tabular learning~\citep{harasymiv2015lessons}. 
There have been many attempts to improve the performance of gradient boosting models using deep networks.
However, they only achieve similar performance to gradient boosting models although they utilize a large number of computing resources~\citep{miller2017forward,zhou2017deep,ke2018tabnn,lay2018random,yang2018deep,feng2018multi}.
Note that we describe the remainder of this paper using gradient boosting models as representative models of tabular learning since they show better performance than deep neural network-based models.

%% file: Sections/3_solution.tex
\section{Efficient Click-Through Rate Prediction}
\label{sec:3_solution}

Although tabular learning and CTR prediction have developed orthogonal to each other so far, they share similar formulation of problem definition.
\begin{equation}
    \arg \min_{f} \sum_{i} \mathcal{L}(y^i, f(x_1^i,  \dotsb, x_n^i))
\end{equation}
It is to find the function $f$ where minimizes the sum of difference, which is measured by a loss function $\mathcal{L}$ over all the $i$-th data instances between actual target $y^i$ and the output value of function $f$ with given input $x^i$. 
The major difference between tabular learning and CTR prediction is that each $x$ is often assumed to be the set of heterogeneous \textit{numerical} features in tubular learning while it is assumed to be the set of highly sparse categorical features in CTR prediction.
In CTR prediction, for instance, $x_k$ ($k=1, \dotsb, n$) can be the categorical index of the user $u$ or item $v$, the gender or age group of users, and the item category.

Conventional gradient boosting is not capable of handling categorical features. 
To take the edge off this problem, several methods have been proposed to convert categorical features into numerical features such as one-hot encoding, Label Encoding (LE), and Target Encoding (TE), but they are not suitable for the CTR prediction task.
One-hot encoding is not practical because the input dimension becomes too large when we apply one-hot encoding to highly sparse categorical features (having extremely high cardinality).
LE, which converts a categorical feature into an arbitrary number, would show sub-optimal performance since there are few correlations between the target category and its encoded number. 
Meanwhile, TE changes the categorical feature into an informative number by calculating the mean of target values with each categorical feature~\citep{micci2001preprocessing}.
However, TE causes overfitting by giving excessive information on each categorical feature~\citep{schifferer2020gpu}.

Recently, there have been new attempts to process categorical features in tabular learning. 
\citet{ke2017lightgbm} adopts \citet{fisher1958grouping} to find the optimal split over categorical features. 
CatBoost Encoding~\citep{dorogush2018catboost} is a variant of TE preventing overfitting by random permutation.
In addition, K-Fold Target Encoding~\citep{ayria2020kfoldte} is intended to increase generality of TE through the K-fold validation.
\citet{schifferer2020gpu} won a competition by applying the K-fold TE to XGBoost~\citep{chen2015xgboost}.

Recent advances in encoding methods of categorical features make it possible for tabular learning to be used for CTR prediction although they do not focus on categorical features with tremendously high cardinality. 
Nevertheless, most CTR prediction models preclude gradient boosting as a baseline~\citep{qu2018product,song2019autoint,cheng2020adaptive}. 
\citet{he2014practical,juan2016field} employ gradient boosting as not a baseline but a feature pre-processing method.
Only a few studies~\citep{ke2019deepgbm} have compared gradient boosting to their model.
However, they do not take advantage of the recent advance in encoding methods to deal with categorical features.

Consequently, we suggest gradient boosting models with the recent advance in handling categorical features as the baselines of the CTR prediction task. 
In Section~\ref{sec:4_experiments}, we will show that gradient boosting not only can be trained at low cost, but also shows better performance than existing CTR prediction models.
These results demonstrate that gradient boosting is suitable for use in developing countries with limited resources.
Noticeably, it is known that considering high order interaction between features is important in the CTR prediction task and gradient boosting also has the capability of modeling high order interaction by depth ($>1$) of the decision tree. 

%% file: Sections/4_experiments.tex
\input{tables/evaluation_results}
\input{tables/evaluation_ablation}
\input{figures/efficiency}

\input{figures/hk_ctr}

\section{Experiments}\label{sec:4_experiments}

In this paper, we compare gradient boosting models with existing CTR prediction models.
In Section~\ref{subsec:4_1_performance_comparison}, we assess the performance of each model on CTR prediction benchmark datasets.
In Section~\ref{subsec:4_2_efficiency_comparison}, we conduct experiments to show the cost-efficiency of gradient boosting models.
In Section~\ref{subsec:4_3_ablation_test}, we verify how much recent categorical feature encoding methods contribute to the performance through ablation study.
In Section~\ref{subsec:4_4_online_deployment_results}, we demonstrate the performance of gradient boosting models on online experiments and the possibility to alleviate the stale problem at a low cost.

\paragraph{Datasets.}{
To assess the performance in CTR prediction, we conduct experiments on the following eight public CTR prediction datasets: \citet{kdd12dataset}, \citet{criteodataset}, \citet{avazudataset}, \citet{talkingdatadataset}, Movielens~\citep{harper2015movielens}, Book-Crossing~\citep{ziegler2005book_crossing}, and \citet{frappedataset}.
We arbitrarily split each dataset into the train, valid, and test sets in a ratio of 8:1:1\footnote{For KDD12, Criteo, Avazu, and Talking Data, 10\% random sampling is used because they are too large not to be suitable for our extensive experimentation}.
}

\paragraph{Baseline Models.}{

Three gradient boosting models and twelve CTR prediction models are considered to justify the efficiency and effectiveness of gradient boosting models in CTR prediction.
XGBoost~\citep{chen2015xgboost}, LightGBM~\citep{ke2017lightgbm} and CatBoost~\citep{dorogush2018catboost} are considered for gradient boosting models. 
Since the originally proposed XGBoost uses LE, K-fold TE~\citep{ayria2020kfoldte} is applied to XGBoost following~\citet{schifferer2020gpu}.
Considered CTR prediction models are as follows:
FM~\citep{rendle2010factorization}, FFM~\citep{juan2016field}, AFM~\citep{xiao2017attentional}, DCN~\citep{wang2017deep}, MLP, NFM~\citep{he2017neural}, Wide \& Deep~\citep{cheng2016wide}, DeepFM~\citep{guo2017deepfm}, xDeepFM~\citep{lian2018xdeepfm}, PNN~\citep{qu2018product}, AutoInt~\citep{song2019autoint}, and AFN~\citep{cheng2020adaptive}.
In developing countries, since it is difficult to perform hyper-parameter tuning every day, we do not newly tune hyper-parameters and do our best to keep originally hyper-parameters reported in each paper.
}

\subsection{Performance Comparison}\label{subsec:4_1_performance_comparison}

The evaluation results are summarized in Table~\ref{table:evaluation_results}.
CatBoost, which is one of the gradient boosting models, outperforms all the baseline models on all the datasets with a large margin while a slight increase in AUROC or decrease in Logloss at \textit{.001-level} is known as a significant improvement in CTR prediction as pointed out in previous works~\citep{cheng2016wide,guo2017deepfm,song2019autoint}.
In addition, the other gradient boosting methods (XGBoost and LightGBM) achieve better or comparable performance to CTR prediction models over all the benchmark datasets. 
\subsection{Efficiency Comparison}\label{subsec:4_2_efficiency_comparison}

In each dataset, the efficiency of gradient boosting models is validated by plotting the change of AUROC according to the training cost
\footnote{Although we only report AUROC, Logloss shows similar trends.}.
Training costs are estimated on AWS EC2 instances.
\texttt{c5a.4xlarge} is used for gradient boosting models because they do not need GPUs. \texttt{p3.2xlarge} is used for CTR prediction models because they require GPUs.
For gradient boosting models, training cost and AUROC in every ten epochs are also reported.
Boosting models shows dramatic improvement in the performance at the early stages of training.

\subsection{Ablation Study}\label{subsec:4_3_ablation_test}
By replacing the latest categorical feature encoding methods with LE and TE, Table~\ref{table:evaluation_ablation} shows that how much the leverage of recent advance of the categorical encoding methods contributes to performance improvement.
Mostly, recent categorical feature encoding methods show statistically significant better results.

\subsection{Online Experiments}\label{subsec:4_4_online_deployment_results}

\input{tables/ctr_online}

Based on the offline test in Section~\ref{subsec:4_1_performance_comparison}, CatBoost is adopted for deploying to our online application (\textit{Hakuna Live}\footnote{\url{https://hakuna.live/}}) downloaded more than 20M times.
Table~\ref{table:ctr_online} shows CTR gain by CatBoost relative to the control group (heuristic algorithm) in online A/B test for seven days on two main regions (Region X and Y). 
CatBoost outperforms the existing heuristic algorithm with a large margin.
Not only that, after we deployed our first model, we experienced a stale problem that performance continued to decline.
To solve this problem, we started daily training, and as a result, we were able to solve the stale problem (See Figure~\ref{fig:online_ctr}).

%% file: tables/evaluation_results.tex
\begin{table*}[t]
\centering
\caption{Evaluation results of three tabular learning models and twelve CTR prediction models on eight real-world datasets. 
Logloss and AUROC with 95\% confidence
interval of 10-runs is provided.
}

\label{table:evaluation_results}

\resizebox{1.0\textwidth}{!}{%

\begin{tabular}{cccccccccc}
\toprule
\multirow{1}{*}{} & \multirow{1}{*}{Model}
& \multicolumn{1}{c}{KDD12}
& \multicolumn{1}{c}{Criteo}
& \multicolumn{1}{c}{Avazu}
& \multicolumn{1}{c}{Talking Data}
& \multicolumn{1}{c}{Amazon}
& \multicolumn{1}{c}{Movielens}
& \multicolumn{1}{c}{Book Crossing}
& \multicolumn{1}{c}{Frappe}
\\
\midrule
\multirow{15}{*}{\STAB{\rotatebox[origin=c]{90}{Logloss}}} & \multirow{1}{*}{XGBoost} & 0.1588 $\pm$ 0.0001 & 0.4595 $\pm$ 0.0050 & 0.3901 $\pm$ 0.0014 & 0.1327 $\pm$ 0.0008 & 0.4947 $\pm$ 0.0003 & 0.2831 $\pm$ 0.0011 & 0.5141 $\pm$ 0.0001 & 0.2849 $\pm$ 0.0019 \\

& \multirow{1}{*}{LightGBM} & 0.1602 $\pm$ 0.0000 & 0.4569 $\pm$ 0.0003 & 0.3916 $\pm$ 0.0003 & 0.1319 $\pm$ 0.0003 & 0.5627 $\pm$ 0.0000 & 0.2437 $\pm$ 0.0014 & 0.5191 $\pm$ 0.0006 & 0.1176 $\pm$ 0.0022 \\

& \multirow{1}{*}{CatBoost} & \textbf{0.1584 $\pm$ 0.0001} & \textbf{0.4507 $\pm$ 0.0002} & \textbf{0.3840 $\pm$ 0.0001} & \textbf{0.1284 $\pm$ 0.0001} & \textbf{0.2221 $\pm$ 0.0004} & \textbf{0.1192 $\pm$ 0.0003} & \textbf{0.4962 $\pm$ 0.0001} & \textbf{0.0780 $\pm$ 0.0005} \\
\cmidrule{2-10}
& \multirow{1}{*}{FM} & 0.1595 $\pm$ 0.0001 & 0.4575 $\pm$ 0.0002 & 0.3912 $\pm$ 0.0004 & 0.1342 $\pm$ 0.0008 & 0.5257 $\pm$ 0.0036 & 0.2783 $\pm$ 0.0027 & 0.5224 $\pm$ 0.0009 & 0.2125 $\pm$ 0.0053 \\

& \multirow{1}{*}{FFM} & 0.1599 $\pm$ 0.0001 & 0.4522 $\pm$ 0.0001 & 0.3899 $\pm$ 0.0002 & 0.1347 $\pm$ 0.0003 & 0.4780 $\pm$ 0.0007 & 0.2414 $\pm$ 0.0040 & 0.5143 $\pm$ 0.0008 & 0.1651 $\pm$ 0.0020 \\

& \multirow{1}{*}{AFM} & 0.1607 $\pm$ 0.0005 & 0.4605 $\pm$ 0.0005 & 0.3941 $\pm$ 0.0002 & 0.1389 $\pm$ 0.0020 & 0.5498 $\pm$ 0.0063 & 0.2714 $\pm$ 0.0112 & 0.5229 $\pm$ 0.0045 & 0.2648 $\pm$ 0.0051 \\

& \multirow{1}{*}{DCN} & 0.1599 $\pm$ 0.0001 & 0.4596 $\pm$ 0.0002 & 0.3926 $\pm$ 0.0003 & 0.1351 $\pm$ 0.0005 & 0.4304 $\pm$ 0.0021 & 0.2897 $\pm$ 0.0006 & 0.5226 $\pm$ 0.0011 & 0.2400 $\pm$ 0.0066 \\

& \multirow{1}{*}{NFM} & 0.1626 $\pm$ 0.0015 & 0.4568 $\pm$ 0.0005 & 0.3910 $\pm$ 0.0005 & 0.1338 $\pm$ 0.0004 & 0.4922 $\pm$ 0.0153 & 0.2862 $\pm$ 0.0243 & 0.5256 $\pm$ 0.0016 & 0.1418 $\pm$ 0.0074 \\

& \multirow{1}{*}{MLP} & 0.1593 $\pm$ 0.0001 & 0.4568 $\pm$ 0.0003 & 0.3923 $\pm$ 0.0006 & 0.1334 $\pm$ 0.0008 & 0.4228 $\pm$ 0.0018 & 0.2829 $\pm$ 0.0068 & 0.5233 $\pm$ 0.0012 & 0.2375 $\pm$ 0.0152 \\

& \multirow{1}{*}{Wide \& Deep} & 0.1593 $\pm$ 0.0001 & 0.4567 $\pm$ 0.0003 & 0.3921 $\pm$ 0.0005 & 0.1335 $\pm$ 0.0009 & 0.4273 $\pm$ 0.0013 & 0.2833 $\pm$ 0.0044 & 0.5232 $\pm$ 0.0011 & 0.2387 $\pm$ 0.0274 \\

& \multirow{1}{*}{DeepFM} & 0.1602 $\pm$ 0.0001 & 0.4586 $\pm$ 0.0004 & 0.3920 $\pm$ 0.0004 & 0.1343 $\pm$ 0.0006 & 0.4317 $\pm$ 0.0024 & 0.2889 $\pm$ 0.0020 & 0.5219 $\pm$ 0.0019 & 0.2244 $\pm$ 0.0102 \\

& \multirow{1}{*}{xDeepFM} & 0.1603 $\pm$ 0.0002 & 0.4589 $\pm$ 0.0004 & 0.3926 $\pm$ 0.0004 & 0.1344 $\pm$ 0.0006 & 0.4346 $\pm$ 0.0017 & 0.2883 $\pm$ 0.0012 & 0.5223 $\pm$ 0.0018 & 0.2214 $\pm$ 0.0142 \\

& \multirow{1}{*}{PNN} & 0.1604 $\pm$ 0.0001 & 0.4573 $\pm$ 0.0003 & 0.3927 $\pm$ 0.0003 & 0.1357 $\pm$ 0.0005 & 0.4355 $\pm$ 0.0028 & 0.2902 $\pm$ 0.0013 & 0.5225 $\pm$ 0.0007 & 0.1957 $\pm$ 0.0057 \\

& \multirow{1}{*}{AutoInt} & 0.1611 $\pm$ 0.0001 & 0.4615 $\pm$ 0.0002 & 0.3953 $\pm$ 0.0006 & 0.1409 $\pm$ 0.0014 & 0.4556 $\pm$ 0.0043 & 0.2906 $\pm$ 0.0011 & 0.5236 $\pm$ 0.0014 & 0.2736 $\pm$ 0.0233 \\

& \multirow{1}{*}{AFN} & 0.1598 $\pm$ 0.0002 & 0.4552 $\pm$ 0.0006 & 0.3912 $\pm$ 0.0009 & 0.1327 $\pm$ 0.0005 & 0.4274 $\pm$ 0.0016 & 0.2768 $\pm$ 0.0036 & 0.5259 $\pm$ 0.0024 & 0.1853 $\pm$ 0.0046 \\
\midrule
\multirow{15}{*}{\STAB{\rotatebox[origin=c]{90}{AUROC}}} & \multirow{1}{*}{XGBoost} & 0.7646 $\pm$ 0.0003 & 0.7889 $\pm$ 0.0061 & 0.7613 $\pm$ 0.0028 & 0.9719 $\pm$ 0.0003 & 0.7395 $\pm$ 0.0005 & 0.9380 $\pm$ 0.0006 & 0.8019 $\pm$ 0.0002 & 0.9353 $\pm$ 0.0007 \\

& \multirow{1}{*}{LightGBM} & 0.7572 $\pm$ 0.0003 & 0.7922 $\pm$ 0.0003 & 0.7584 $\pm$ 0.0006 & 0.9724 $\pm$ 0.0001 & 0.4946 $\pm$ 0.0009 & 0.9483 $\pm$ 0.0008 & 0.7951 $\pm$ 0.0006 & 0.9855 $\pm$ 0.0005 \\

& \multirow{1}{*}{CatBoost} & \textbf{0.7655 $\pm$ 0.0003} & \textbf{0.7993 $\pm$ 0.0002} & \textbf{0.7734 $\pm$ 0.0002} & \textbf{0.9736 $\pm$ 0.0001} & \textbf{0.9598 $\pm$ 0.0001} & \textbf{0.9848 $\pm$ 0.0001} & \textbf{0.8183 $\pm$ 0.0001} & \textbf{0.9925 $\pm$ 0.0002} \\
\cmidrule{2-10}
& \multirow{1}{*}{FM} & 0.7618 $\pm$ 0.0006 & 0.7922 $\pm$ 0.0003 & 0.7600 $\pm$ 0.0008 & 0.9710 $\pm$ 0.0003 & 0.6823 $\pm$ 0.0118 & 0.9375 $\pm$ 0.0014 & 0.7945 $\pm$ 0.0009 & 0.9659 $\pm$ 0.0012 \\

& \multirow{1}{*}{FFM} & 0.7610 $\pm$ 0.0004 & 0.7979 $\pm$ 0.0001 & 0.7622 $\pm$ 0.0005 & 0.9707 $\pm$ 0.0002 & 0.7592 $\pm$ 0.0008 & 0.9529 $\pm$ 0.0015 & 0.8014 $\pm$ 0.0007 & 0.9752 $\pm$ 0.0004 \\

& \multirow{1}{*}{AFM} & 0.7571 $\pm$ 0.0029 & 0.7883 $\pm$ 0.0006 & 0.7541 $\pm$ 0.0005 & 0.9689 $\pm$ 0.0010 & 0.6013 $\pm$ 0.0334 & 0.9423 $\pm$ 0.0077 & 0.7938 $\pm$ 0.0037 & 0.9544 $\pm$ 0.0014 \\

& \multirow{1}{*}{DCN} & 0.7602 $\pm$ 0.0006 & 0.7895 $\pm$ 0.0005 & 0.7569 $\pm$ 0.0006 & 0.9700 $\pm$ 0.0004 & 0.8146 $\pm$ 0.0018 & 0.9348 $\pm$ 0.0004 & 0.7929 $\pm$ 0.0008 & 0.9669 $\pm$ 0.0014 \\

& \multirow{1}{*}{NFM} & 0.7576 $\pm$ 0.0014 & 0.7929 $\pm$ 0.0007 & 0.7600 $\pm$ 0.0006 & 0.9712 $\pm$ 0.0004 & 0.7402 $\pm$ 0.0212 & 0.9441 $\pm$ 0.0047 & 0.7917 $\pm$ 0.0014 & 0.9821 $\pm$ 0.0011 \\

& \multirow{1}{*}{MLP} & 0.7627 $\pm$ 0.0005 & 0.7929 $\pm$ 0.0005 & 0.7574 $\pm$ 0.0012 & 0.9717 $\pm$ 0.0003 & 0.8233 $\pm$ 0.0013 & 0.9413 $\pm$ 0.0036 & 0.7926 $\pm$ 0.0012 & 0.9628 $\pm$ 0.0055 \\

& \multirow{1}{*}{Wide \& Deep} & 0.7626 $\pm$ 0.0004 & 0.7930 $\pm$ 0.0005 & 0.7580 $\pm$ 0.0005 & 0.9715 $\pm$ 0.0010 & 0.8184 $\pm$ 0.0009 & 0.9407 $\pm$ 0.0033 & 0.7932 $\pm$ 0.0025 & 0.9648 $\pm$ 0.0034 \\

& \multirow{1}{*}{DeepFM} & 0.7585 $\pm$ 0.0005 & 0.7908 $\pm$ 0.0006 & 0.7581 $\pm$ 0.0006 & 0.9705 $\pm$ 0.0006 & 0.8139 $\pm$ 0.0024 & 0.9354 $\pm$ 0.0018 & 0.7944 $\pm$ 0.0008 & 0.9630 $\pm$ 0.0035 \\

& \multirow{1}{*}{xDeepFM} & 0.7580 $\pm$ 0.0007 & 0.7904 $\pm$ 0.0005 & 0.7572 $\pm$ 0.0006 & 0.9705 $\pm$ 0.0004 & 0.8101 $\pm$ 0.0025 & 0.9366 $\pm$ 0.0019 & 0.7941 $\pm$ 0.0013 & 0.9686 $\pm$ 0.0038 \\

& \multirow{1}{*}{PNN} & 0.7584 $\pm$ 0.0007 & 0.7927 $\pm$ 0.0006 & 0.7569 $\pm$ 0.0004 & 0.9702 $\pm$ 0.0002 & 0.8094 $\pm$ 0.0029 & 0.9361 $\pm$ 0.0013 & 0.7934 $\pm$ 0.0008 & 0.9742 $\pm$ 0.0015 \\

& \multirow{1}{*}{AutoInt} & 0.7559 $\pm$ 0.0018 & 0.7872 $\pm$ 0.0002 & 0.7516 $\pm$ 0.0010 & 0.9682 $\pm$ 0.0006 & 0.7873 $\pm$ 0.0055 & 0.9361 $\pm$ 0.0004 & 0.7927 $\pm$ 0.0009 & 0.9427 $\pm$ 0.0117 \\

& \multirow{1}{*}{AFN} & 0.7598 $\pm$ 0.0016 & 0.7947 $\pm$ 0.0007 & 0.7600 $\pm$ 0.0015 & 0.9717 $\pm$ 0.0005 & 0.8173 $\pm$ 0.0017 & 0.9430 $\pm$ 0.0016 & 0.7921 $\pm$ 0.0025 & 0.9760 $\pm$ 0.0018 \\
\bottomrule
\end{tabular}

}
\vspace{-1\baselineskip}
\end{table*}

%% file: tables/evaluation_ablation.tex
\begin{table*}[t]
\caption{Ablation study results of three tabular learning models regarding to encoding methods of categorical features. Logloss and AUROC with 95\% confidence interval of 10-runs is provided.}

\label{table:evaluation_ablation}

\resizebox{1.0\textwidth}{!}{%

\begin{tabular}{ccccccccccc}
\toprule
 & \multirow{1}{*}{Model} & \multirow{1}{*}{Encoding}
& \multicolumn{1}{c}{KDD12}
& \multicolumn{1}{c}{Criteo}
& \multicolumn{1}{c}{Avazu}
& \multicolumn{1}{c}{Talking Data}
& \multicolumn{1}{c}{Amazon}
& \multicolumn{1}{c}{Movielens}
& \multicolumn{1}{c}{Book Crossing}
& \multicolumn{1}{c}{Frappe}
\\
\midrule
\multirow{9}{*}{\STAB{\rotatebox[origin=c]{90}{Logloss}}} & \multirow{3}{*}{XGBoost} & LE & 0.1624 $\pm$ 0.0002 & \textbf{0.4613 $\pm$ 0.0009} & 0.3930 $\pm$ 0.0007 & \textbf{0.1332 $\pm$ 0.0007} & 0.5532 $\pm$ 0.0010 & 0.3125 $\pm$ 0.0012 & 0.5415 $\pm$ 0.0033 & \textbf{0.1805 $\pm$ 0.0098} \\

& & TE & 0.1723 $\pm$ 0.0010 & 0.4726 $\pm$ 0.0008 & 0.5194 $\pm$ 0.0025 & 0.1543 $\pm$ 0.0023 & 0.6047 $\pm$ 0.0004 & 0.3619 $\pm$ 0.0014 & 0.8306 $\pm$ 0.0006 & 0.3059 $\pm$ 0.0011 \\

& & - & \textbf{0.1588 $\pm$ 0.0001} & \textbf{0.4595 $\pm$ 0.0050} & \textbf{0.3901 $\pm$ 0.0014} & \textbf{0.1327 $\pm$ 0.0008} & \textbf{0.4947 $\pm$ 0.0003} & \textbf{0.2831 $\pm$ 0.0011} & \textbf{0.5141 $\pm$ 0.0001} & 0.2849 $\pm$ 0.0019 \\
\cmidrule{2-11}
 & \multirow{3}{*}{LightGBM} & LE & 0.1617 $\pm$ 0.0004 & 0.4628 $\pm$ 0.0010 & \textbf{0.3913 $\pm$ 0.0005} & \textbf{0.1302 $\pm$ 0.0002} & \textbf{0.5151 $\pm$ 0.0006} & 0.4609 $\pm$ 0.0005 & 0.5383 $\pm$ 0.0049 & 0.2576 $\pm$ 0.0068 \\

& & TE & 0.1656 $\pm$ 0.0001 & 0.4710 $\pm$ 0.0001 & 0.4260 $\pm$ 0.0001 & 0.1439 $\pm$ 0.0003 & 0.5661 $\pm$ 0.0001 & 0.3226 $\pm$ 0.0002 & 0.6013 $\pm$ 0.0004 & 0.2942 $\pm$ 0.0003 \\

& & - & \textbf{0.1602 $\pm$ 0.0000} & \textbf{0.4569 $\pm$ 0.0003} & \textbf{0.3916 $\pm$ 0.0003} & 0.1319 $\pm$ 0.0003 & 0.5627 $\pm$ 0.0000 & \textbf{0.2437 $\pm$ 0.0014} & \textbf{0.5191 $\pm$ 0.0006} & \textbf{0.1176 $\pm$ 0.0022} \\
\cmidrule{2-11}
 & \multirow{3}{*}{CatBoost} & LE & 0.1622 $\pm$ 0.0001 & 0.4656 $\pm$ 0.0004 & 0.3924 $\pm$ 0.0001 & 0.1303 $\pm$ 0.0001 & 0.5232 $\pm$ 0.0003 & 0.4776 $\pm$ 0.0009 & 0.5507 $\pm$ 0.0002 & 0.2888 $\pm$ 0.0014 \\

& & TE & 0.1683 $\pm$ 0.0032 & 0.4654 $\pm$ 0.0005 & 0.4774 $\pm$ 0.0226 & 0.1447 $\pm$ 0.0011 & 0.5943 $\pm$ 0.0024 & 0.3209 $\pm$ 0.0100 & 0.6252 $\pm$ 0.0143 & 0.2795 $\pm$ 0.0021 \\

& & - & \textbf{0.1584 $\pm$ 0.0001} & \textbf{0.4507 $\pm$ 0.0002} & \textbf{0.3840 $\pm$ 0.0001} & \textbf{0.1284 $\pm$ 0.0001} & \textbf{0.2221 $\pm$ 0.0004} & \textbf{0.1192 $\pm$ 0.0003} & \textbf{0.4962 $\pm$ 0.0001} & \textbf{0.0780 $\pm$ 0.0005} \\
\midrule
\multirow{9}{*}{\STAB{\rotatebox[origin=c]{90}{AUROC}}} & \multirow{3}{*}{XGBoost} & LE & 0.7422 $\pm$ 0.0013 & \textbf{0.7869 $\pm$ 0.0011} & 0.7564 $\pm$ 0.0013 & \textbf{0.9718 $\pm$ 0.0003} & 0.5833 $\pm$ 0.0042 & 0.9241 $\pm$ 0.0006 & 0.7743 $\pm$ 0.0032 & \textbf{0.9721 $\pm$ 0.0025} \\

& & TE & 0.7258 $\pm$ 0.0012 & 0.7768 $\pm$ 0.0009 & 0.6960 $\pm$ 0.0004 & 0.9572 $\pm$ 0.0021 & 0.3054 $\pm$ 0.0032 & 0.9043 $\pm$ 0.0008 & 0.7029 $\pm$ 0.0010 & 0.9295 $\pm$ 0.0005 \\

& & - & \textbf{0.7646 $\pm$ 0.0003} & \textbf{0.7889 $\pm$ 0.0061} & \textbf{0.7613 $\pm$ 0.0028} & \textbf{0.9719 $\pm$ 0.0003} & \textbf{0.7395 $\pm$ 0.0005} & \textbf{0.9380 $\pm$ 0.0006} & \textbf{0.8019 $\pm$ 0.0002} & 0.9353 $\pm$ 0.0007 \\
\cmidrule{2-11}
 & \multirow{3}{*}{LightGBM} & LE & 0.7468 $\pm$ 0.0021 & 0.7852 $\pm$ 0.0013 & \textbf{0.7599 $\pm$ 0.0009} & \textbf{0.9730 $\pm$ 0.0001} & \textbf{0.6892 $\pm$ 0.0012} & 0.8324 $\pm$ 0.0005 & 0.7793 $\pm$ 0.0054 & 0.9514 $\pm$ 0.0026 \\

& & TE & 0.7165 $\pm$ 0.0007 & 0.7767 $\pm$ 0.0006 & 0.6939 $\pm$ 0.0003 & 0.9652 $\pm$ 0.0003 & 0.2908 $\pm$ 0.0092 & 0.9091 $\pm$ 0.0001 & 0.6967 $\pm$ 0.0059 & 0.9290 $\pm$ 0.0003 \\

& & - & \textbf{0.7572 $\pm$ 0.0003} & \textbf{0.7922 $\pm$ 0.0003} & \textbf{0.7584 $\pm$ 0.0006} & 0.9724 $\pm$ 0.0001 & 0.4946 $\pm$ 0.0009 & \textbf{0.9483 $\pm$ 0.0008} & \textbf{0.7951 $\pm$ 0.0006} & \textbf{0.9855 $\pm$ 0.0005} \\
\cmidrule{2-11}
 & \multirow{3}{*}{CatBoost} & LE & 0.7436 $\pm$ 0.0003 & 0.7816 $\pm$ 0.0005 & 0.7579 $\pm$ 0.0003 & 0.9729 $\pm$ 0.0001 & 0.6711 $\pm$ 0.0009 & 0.8164 $\pm$ 0.0008 & 0.7641 $\pm$ 0.0004 & 0.9360 $\pm$ 0.0006 \\

& & TE & 0.7300 $\pm$ 0.0065 & 0.7867 $\pm$ 0.0012 & 0.6822 $\pm$ 0.0179 & 0.9654 $\pm$ 0.0005 & 0.2798 $\pm$ 0.0069 & 0.9086 $\pm$ 0.0013 & 0.7047 $\pm$ 0.0064 & 0.9395 $\pm$ 0.0008 \\

& & - & \textbf{0.7655 $\pm$ 0.0003} & \textbf{0.7993 $\pm$ 0.0002} & \textbf{0.7734 $\pm$ 0.0002} & \textbf{0.9736 $\pm$ 0.0001} & \textbf{0.9598 $\pm$ 0.0001} & \textbf{0.9848 $\pm$ 0.0001} & \textbf{0.8183 $\pm$ 0.0001} & \textbf{0.9925 $\pm$ 0.0002} \\
\bottomrule
\end{tabular}

}
\vspace{-1\baselineskip}
\end{table*}

%% file: figures/efficiency.tex
\begin{figure*}[t] 
\begin{center}
\includegraphics[width=0.96\textwidth]{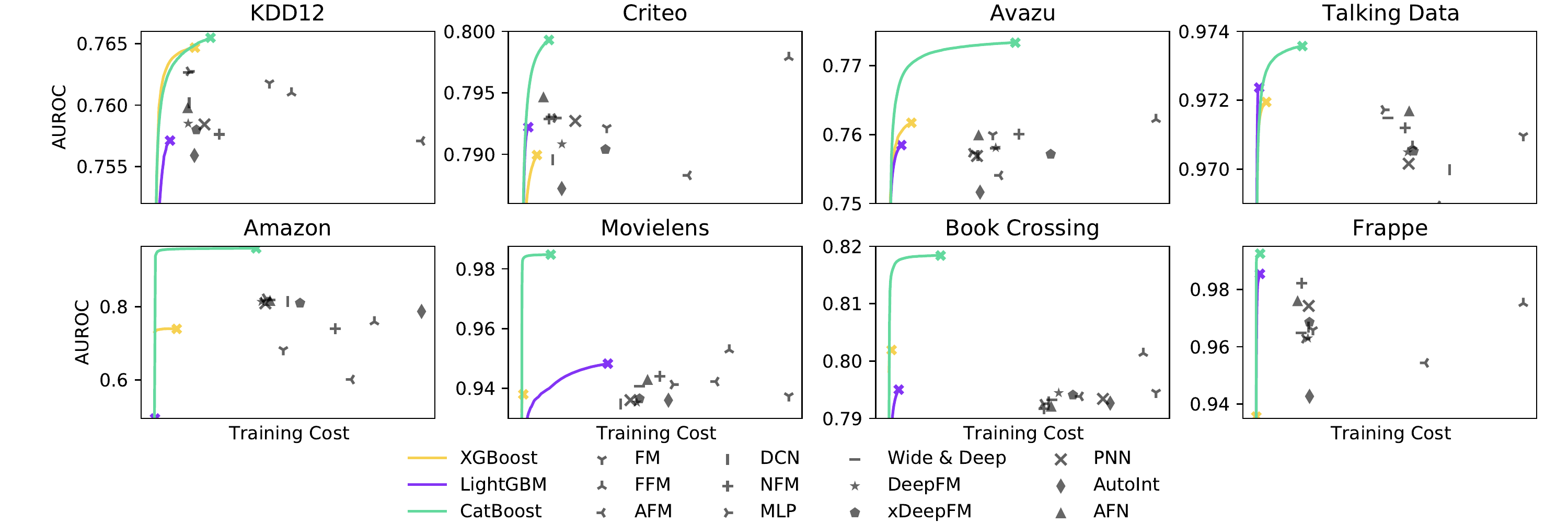}
\end{center}
\caption{AUROC by training cost estimated on AWS EC2 instances.}
\label{fig:efficiency}
\end{figure*}

%% file: figures/hk_ctr.tex
\begin{figure*}[t] 
\begin{center}
\includegraphics[width=0.9\textwidth]{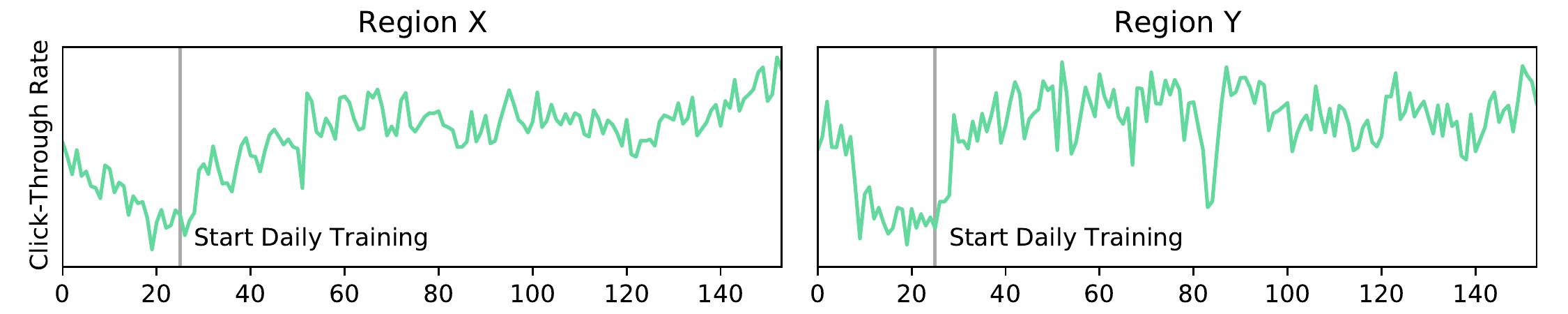}
\end{center}
\caption{
Alleviating stale problem by daily training with CatBoost.
CTR over time after first model deployment on two main regions X and Y of our application is plotted. 
}
\label{fig:online_ctr}
\end{figure*}

%% file: tables/ctr_online.tex
\begin{wraptable}{R}{0.24\textwidth}
\begin{minipage}{0.24\textwidth}
\vspace{-0.16in}
\caption{CTR gain of CatBoost model in online A/B test.}
\label{table:ctr_online}
\begin{center}

\resizebox{1.0\columnwidth}{!}{%
\begin{tabular}{lrrr}
\toprule
& CTR Gain \\
\midrule 
Region X & + 59.47\% \\
Region Y & + 84.96\% \\
\bottomrule
\end{tabular}
}

\end{center}
\end{minipage}
\end{wraptable}

%% file: Sections/5_conclusion.tex
\section{Conclusion}

We suggest tabular learning models as CTR prediction for developing countries by explicitly shedding some light on the relationship between tabular learning and CTR prediction task. 
The state-of-the-art performance on eight public datasets and better results on online A/B test can be achieved at a low cost with tabular learning models (especially gradient boosting) and the recent advance in categorical feature encoding methods. 
In addition, our study provides room for improvement of applications in developing countries under limited computing resources by showing that the state-of-the-art performance is achieved with tabular learning models with the recent advance in categorical feature encoding methods.

%% file: iclr2020_conference.bbl
\begin{thebibliography}{45}
\providecommand{\natexlab}[1]{#1}
\providecommand{\url}[1]{\texttt{#1}}
\expandafter\ifx\csname urlstyle\endcsname\relax
  \providecommand{\doi}[1]{doi: #1}\else
  \providecommand{\doi}{doi: \begingroup \urlstyle{rm}\Url}\fi

\bibitem[Avazu()]{avazudataset}
Avazu.
\newblock Avazu dataset.
\newblock \url{https://www.kaggle.com/c/avazu-ctr-prediction}.
\newblock [Online; accessed 06-March-2020].

\bibitem[Ayria(2020)]{ayria2020kfoldte}
Pourya Ayria.
\newblock K-fold target encoding.
\newblock
  \url{https://medium.com/@pouryaayria/k-fold-target-encoding-dfe9a594874b},
  2020.
\newblock [Online; accessed 06-March-2020].

\bibitem[Chen et~al.(2015)Chen, He, Benesty, Khotilovich, Tang, Cho,
  et~al.]{chen2015xgboost}
Tianqi Chen, Tong He, Michael Benesty, Vadim Khotilovich, Yuan Tang, Hyunsu
  Cho, et~al.
\newblock Xgboost: extreme gradient boosting.
\newblock \emph{R package version 0.4-2}, 1\penalty0 (4), 2015.

\bibitem[Chen et~al.(2019)Chen, Chen, He, Gao, Li, Lou, and
  Wang]{chen2019lambdaopt}
Yihong Chen, Bei Chen, Xiangnan He, Chen Gao, Yong Li, Jian-Guang Lou, and Yue
  Wang.
\newblock $\lambda$opt: Learn to regularize recommender models in finer levels.
\newblock In \emph{Proceedings of the 25th ACM SIGKDD International Conference
  on Knowledge Discovery \& Data Mining}, pp.\  978--986, 2019.

\bibitem[Cheng et~al.(2016)Cheng, Koc, Harmsen, Shaked, Chandra, Aradhye,
  Anderson, Corrado, Chai, Ispir, et~al.]{cheng2016wide}
Heng-Tze Cheng, Levent Koc, Jeremiah Harmsen, Tal Shaked, Tushar Chandra,
  Hrishi Aradhye, Glen Anderson, Greg Corrado, Wei Chai, Mustafa Ispir, et~al.
\newblock Wide \& deep learning for recommender systems.
\newblock In \emph{Proceedings of the 1st workshop on deep learning for
  recommender systems}, pp.\  7--10, 2016.

\bibitem[Cheng et~al.(2020)Cheng, Shen, and Huang]{cheng2020adaptive}
Weiyu Cheng, Yanyan Shen, and Linpeng Huang.
\newblock Adaptive factorization network: Learning adaptive-order feature
  interactions.
\newblock In \emph{The Thirty-Fourth {AAAI} Conference on Artificial
  Intelligence, 2020}, pp.\  3609--3616. {AAAI} Press, 2020.
\newblock URL \url{https://aaai.org/ojs/index.php/AAAI/article/view/5768}.

\bibitem[Criteo()]{criteodataset}
Criteo.
\newblock Criteo dataset.
\newblock \url{https://www.kaggle.com/c/criteo-display-ad-challenge}.
\newblock [Online; accessed 06-March-2020].

\bibitem[Data()]{talkingdatadataset}
Talking Data.
\newblock Talking data dataset.
\newblock
  \url{https://www.kaggle.com/c/talkingdata-adtracking-fraud-detection}.
\newblock [Online; accessed 06-March-2020].

\bibitem[Dorogush et~al.(2018)Dorogush, Ershov, and
  Gulin]{dorogush2018catboost}
Anna~Veronika Dorogush, Vasily Ershov, and Andrey Gulin.
\newblock Catboost: gradient boosting with categorical features support.
\newblock \emph{arXiv preprint arXiv:1810.11363}, 2018.

\bibitem[Feng et~al.(2018)Feng, Yu, and Zhou]{feng2018multi}
Ji~Feng, Yang Yu, and Zhi-Hua Zhou.
\newblock Multi-layered gradient boosting decision trees.
\newblock \emph{arXiv preprint arXiv:1806.00007}, 2018.

\bibitem[Fisher(1958)]{fisher1958grouping}
Walter~D Fisher.
\newblock On grouping for maximum homogeneity.
\newblock \emph{Journal of the American statistical Association}, 53\penalty0
  (284):\penalty0 789--798, 1958.

\bibitem[Frappe()]{frappedataset}
Frappe.
\newblock Frappe dataset.
\newblock \url{https://www.baltrunas.info/context-aware}.
\newblock [Online; accessed 06-March-2020].

\bibitem[Guo et~al.(2017)Guo, Tang, Ye, Li, and He]{guo2017deepfm}
Huifeng Guo, Ruiming Tang, Yunming Ye, Zhenguo Li, and Xiuqiang He.
\newblock Deepfm: A factorization-machine based neural network for ctr
  prediction.
\newblock In \emph{Proceedings of the 26th International Joint Conference on
  Artificial Intelligence}, IJCAI'17, pp.\  1725–1731. AAAI Press, 2017.
\newblock ISBN 9780999241103.

\bibitem[Harasymiv(2015)]{harasymiv2015lessons}
Vasyl Harasymiv.
\newblock Lessons from 2 million machine learning models on kaggle.
\newblock
  \url{https://www.kdnuggets.com/2015/12/harasymiv-lessons-kaggle-machine-learning.html},
  2015.
\newblock [Online; accessed 06-March-2020].

\bibitem[Harper \& Konstan(2015)Harper and Konstan]{harper2015movielens}
F~Maxwell Harper and Joseph~A Konstan.
\newblock The movielens datasets: History and context.
\newblock \emph{Acm transactions on interactive intelligent systems (tiis)},
  5\penalty0 (4):\penalty0 1--19, 2015.

\bibitem[He \& Chua(2017)He and Chua]{he2017neural}
Xiangnan He and Tat-Seng Chua.
\newblock Neural factorization machines for sparse predictive analytics.
\newblock In \emph{Proceedings of the 40th International ACM SIGIR conference
  on Research and Development in Information Retrieval}, pp.\  355--364, 2017.

\bibitem[He et~al.(2014)He, Pan, Jin, Xu, Liu, Xu, Shi, Atallah, Herbrich,
  Bowers, et~al.]{he2014practical}
Xinran He, Junfeng Pan, Ou~Jin, Tianbing Xu, Bo~Liu, Tao Xu, Yanxin Shi,
  Antoine Atallah, Ralf Herbrich, Stuart Bowers, et~al.
\newblock Practical lessons from predicting clicks on ads at facebook.
\newblock In \emph{Proceedings of the Eighth International Workshop on Data
  Mining for Online Advertising}, pp.\  1--9, 2014.

\bibitem[Jagerman et~al.(2019)Jagerman, Markov, and
  de~Rijke]{jagerman2019people}
Rolf Jagerman, Ilya Markov, and Maarten de~Rijke.
\newblock When people change their mind: Off-policy evaluation in
  non-stationary recommendation environments.
\newblock In \emph{Proceedings of the Twelfth ACM International Conference on
  Web Search and Data Mining}, pp.\  447--455, 2019.

\bibitem[Juan et~al.(2016)Juan, Zhuang, Chin, and Lin]{juan2016field}
Yuchin Juan, Yong Zhuang, Wei-Sheng Chin, and Chih-Jen Lin.
\newblock Field-aware factorization machines for ctr prediction.
\newblock In \emph{Proceedings of the 10th ACM Conference on Recommender
  Systems}, pp.\  43--50, 2016.

\bibitem[KDD12()]{kdd12dataset}
KDD12.
\newblock Kdd cup 2012, track 2 dataset.
\newblock \url{https://www.kaggle.com/c/kddcup2012-track2}.
\newblock [Online; accessed 06-March-2020].

\bibitem[Ke et~al.(2017)Ke, Meng, Finley, Wang, Chen, Ma, Ye, and
  Liu]{ke2017lightgbm}
Guolin Ke, Qi~Meng, Thomas Finley, Taifeng Wang, Wei Chen, Weidong Ma, Qiwei
  Ye, and Tie-Yan Liu.
\newblock Lightgbm: A highly efficient gradient boosting decision tree.
\newblock \emph{Advances in neural information processing systems},
  30:\penalty0 3146--3154, 2017.

\bibitem[Ke et~al.(2018)Ke, Zhang, Xu, Bian, and Liu]{ke2018tabnn}
Guolin Ke, Jia Zhang, Zhenhui Xu, Jiang Bian, and Tie-Yan Liu.
\newblock Tabnn: A universal neural network solution for tabular data.
\newblock 2018.

\bibitem[Ke et~al.(2019)Ke, Xu, Zhang, Bian, and Liu]{ke2019deepgbm}
Guolin Ke, Zhenhui Xu, Jia Zhang, Jiang Bian, and Tie-Yan Liu.
\newblock Deepgbm: A deep learning framework distilled by gbdt for online
  prediction tasks.
\newblock In \emph{Proceedings of the 25th ACM SIGKDD International Conference
  on Knowledge Discovery \& Data Mining}, pp.\  384--394, 2019.

\bibitem[Koren(2009)]{koren2009collaborative}
Yehuda Koren.
\newblock Collaborative filtering with temporal dynamics.
\newblock In \emph{Proceedings of the 15th ACM SIGKDD international conference
  on Knowledge discovery and data mining}, pp.\  447--456, 2009.

\bibitem[Lay et~al.(2018)Lay, Harrison, Schreiber, Dawer, and
  Barbu]{lay2018random}
Nathan Lay, Adam~P Harrison, Sharon Schreiber, Gitesh Dawer, and Adrian Barbu.
\newblock Random hinge forest for differentiable learning.
\newblock \emph{arXiv preprint arXiv:1802.03882}, 2018.

\bibitem[Lee et~al.(2019)Lee, Im, Jang, Cho, and Chung]{lee2019melu}
Hoyeop Lee, Jinbae Im, Seongwon Jang, Hyunsouk Cho, and Sehee Chung.
\newblock Melu: meta-learned user preference estimator for cold-start
  recommendation.
\newblock In \emph{Proceedings of the 25th ACM SIGKDD International Conference
  on Knowledge Discovery \& Data Mining}, pp.\  1073--1082, 2019.

\bibitem[Lian et~al.(2018)Lian, Zhou, Zhang, Chen, Xie, and
  Sun]{lian2018xdeepfm}
Jianxun Lian, Xiaohuan Zhou, Fuzheng Zhang, Zhongxia Chen, Xing Xie, and
  Guangzhong Sun.
\newblock xdeepfm: Combining explicit and implicit feature interactions for
  recommender systems.
\newblock In \emph{Proceedings of the 24th ACM SIGKDD International Conference
  on Knowledge Discovery \& Data Mining}, pp.\  1754--1763, 2018.

\bibitem[Micci-Barreca(2001)]{micci2001preprocessing}
Daniele Micci-Barreca.
\newblock A preprocessing scheme for high-cardinality categorical attributes in
  classification and prediction problems.
\newblock \emph{ACM SIGKDD Explorations Newsletter}, 3\penalty0 (1):\penalty0
  27--32, 2001.

\bibitem[Miller et~al.(2017)Miller, Hettinger, Humpherys, Jarvis, and
  Kartchner]{miller2017forward}
Kevin Miller, Chris Hettinger, Jeffrey Humpherys, Tyler Jarvis, and David
  Kartchner.
\newblock Forward thinking: Building deep random forests.
\newblock \emph{arXiv preprint arXiv:1705.07366}, 2017.

\bibitem[Nasraoui et~al.(2007)Nasraoui, Cerwinske, Rojas, and
  Gonzalez]{nasraoui2007performance}
Olfa Nasraoui, Jeff Cerwinske, Carlos Rojas, and Fabio Gonzalez.
\newblock Performance of recommendation systems in dynamic streaming
  environments.
\newblock In \emph{Proceedings of the 2007 SIAM International Conference on
  Data Mining}, pp.\  569--574. SIAM, 2007.

\bibitem[Qu et~al.(2018)Qu, Fang, Zhang, Tang, Niu, Guo, Yu, and
  He]{qu2018product}
Yanru Qu, Bohui Fang, Weinan Zhang, Ruiming Tang, Minzhe Niu, Huifeng Guo, Yong
  Yu, and Xiuqiang He.
\newblock Product-based neural networks for user response prediction over
  multi-field categorical data.
\newblock \emph{ACM Transactions on Information Systems (TOIS)}, 37\penalty0
  (1):\penalty0 1--35, 2018.

\bibitem[Radinsky et~al.(2012)Radinsky, Svore, Dumais, Teevan, Bocharov, and
  Horvitz]{radinsky2012modeling}
Kira Radinsky, Krysta Svore, Susan Dumais, Jaime Teevan, Alex Bocharov, and
  Eric Horvitz.
\newblock Modeling and predicting behavioral dynamics on the web.
\newblock In \emph{Proceedings of the 21st international conference on World
  Wide Web}, pp.\  599--608, 2012.

\bibitem[Rendle(2010)]{rendle2010factorization}
Steffen Rendle.
\newblock Factorization machines.
\newblock In \emph{2010 IEEE International Conference on Data Mining}, pp.\
  995--1000. IEEE, 2010.

\bibitem[Rendle et~al.(2012)Rendle, Freudenthaler, Gantner, and
  Schmidt-Thieme]{rendle2012bpr}
Steffen Rendle, Christoph Freudenthaler, Zeno Gantner, and Lars Schmidt-Thieme.
\newblock Bpr: Bayesian personalized ranking from implicit feedback.
\newblock \emph{arXiv preprint arXiv:1205.2618}, 2012.

\bibitem[Schifferer et~al.(2020)Schifferer, Titericz, Deotte, Henkel, Onodera,
  Liu, Tunguz, Oldridge, De~Souza Pereira~Moreira, and
  Erdem]{schifferer2020gpu}
Benedikt Schifferer, Gilberto Titericz, Chris Deotte, Christof Henkel, Kazuki
  Onodera, Jiwei Liu, Bojan Tunguz, Even Oldridge, Gabriel De~Souza
  Pereira~Moreira, and Ahmet Erdem.
\newblock Gpu accelerated feature engineering and training for recommender
  systems.
\newblock In \emph{Proceedings of the Recommender Systems Challenge 2020}, pp.\
   16--23. 2020.

\bibitem[Shi et~al.(2019)Shi, Ozsoy, Hurley, Smyth, Tragos, Geraci, and
  Lawlor]{shi2019pyrecgym}
Bichen Shi, Makbule~Gulcin Ozsoy, Neil Hurley, Barry Smyth, Elias~Z Tragos,
  James Geraci, and Aonghus Lawlor.
\newblock Pyrecgym: a reinforcement learning gym for recommender systems.
\newblock In \emph{Proceedings of the 13th ACM Conference on Recommender
  Systems}, pp.\  491--495, 2019.

\bibitem[Song et~al.(2019)Song, Shi, Xiao, Duan, Xu, Zhang, and
  Tang]{song2019autoint}
Weiping Song, Chence Shi, Zhiping Xiao, Zhijian Duan, Yewen Xu, Ming Zhang, and
  Jian Tang.
\newblock Autoint: Automatic feature interaction learning via self-attentive
  neural networks.
\newblock In \emph{Proceedings of the 28th ACM International Conference on
  Information and Knowledge Management}, pp.\  1161--1170, 2019.

\bibitem[Trevisiol et~al.(2014)Trevisiol, Aiello, Schifanella, and
  Jaimes]{trevisiol2014cold}
Michele Trevisiol, Luca~Maria Aiello, Rossano Schifanella, and Alejandro
  Jaimes.
\newblock Cold-start news recommendation with domain-dependent browse graph.
\newblock In \emph{Proceedings of the 8th ACM Conference on Recommender
  systems}, pp.\  81--88, 2014.

\bibitem[Vartak et~al.(2017)Vartak, Thiagarajan, Miranda, Bratman, and
  Larochelle]{vartak2017meta}
Manasi Vartak, Arvind Thiagarajan, Conrado Miranda, Jeshua Bratman, and Hugo
  Larochelle.
\newblock A meta-learning perspective on cold-start recommendations for items.
\newblock 2017.

\bibitem[Wang et~al.(2019)Wang, Yankov, Evans, Palanisamy, Arora, and
  Wu]{wang2019predicting}
Renzhong Wang, Dragomir Yankov, Michael~R Evans, Senthil Palanisamy, Siddhartha
  Arora, and Wei Wu.
\newblock Predicting user routines with masked dilated convolutions.
\newblock In \emph{Proceedings of the 13th ACM Conference on Recommender
  Systems}, pp.\  481--485, 2019.

\bibitem[Wang et~al.(2017)Wang, Fu, Fu, and Wang]{wang2017deep}
Ruoxi Wang, Bin Fu, Gang Fu, and Mingliang Wang.
\newblock Deep \& cross network for ad click predictions.
\newblock In \emph{Proceedings of the ADKDD'17}, pp.\  1--7. 2017.

\bibitem[Xiao et~al.(2017)Xiao, Ye, He, Zhang, Wu, and
  Chua]{xiao2017attentional}
Jun Xiao, Hao Ye, Xiangnan He, Hanwang Zhang, Fei Wu, and Tat-Seng Chua.
\newblock Attentional factorization machines: learning the weight of feature
  interactions via attention networks.
\newblock In \emph{Proceedings of the 26th International Joint Conference on
  Artificial Intelligence}, pp.\  3119--3125, 2017.

\bibitem[Yang et~al.(2018)Yang, Morillo, and Hospedales]{yang2018deep}
Yongxin Yang, Irene~Garcia Morillo, and Timothy~M Hospedales.
\newblock Deep neural decision trees.
\newblock \emph{arXiv preprint arXiv:1806.06988}, 2018.

\bibitem[Zhou \& Feng(2017)Zhou and Feng]{zhou2017deep}
Zhi-Hua Zhou and Ji~Feng.
\newblock Deep forest.
\newblock \emph{arXiv preprint arXiv:1702.08835}, 2017.

\bibitem[Ziegler et~al.(2005)Ziegler, McNee, Konstan, and
  Lausen]{ziegler2005book_crossing}
Cai-Nicolas Ziegler, Sean~M McNee, Joseph~A Konstan, and Georg Lausen.
\newblock Improving recommendation lists through topic diversification.
\newblock In \emph{Proceedings of the 14th international conference on World
  Wide Web}, pp.\  22--32, 2005.

\end{thebibliography}
